\documentclass{article}
\usepackage{arxiv}
\usepackage{times}

\usepackage{soul}
\usepackage{url}
\usepackage[hidelinks]{hyperref}
\usepackage[utf8]{inputenc}
\usepackage[small]{caption}
\usepackage{graphicx}
\usepackage{amsmath}
\usepackage{booktabs}
\usepackage{tabularx}
\usepackage{listings}
\usepackage{tcolorbox}
\usepackage{xcolor}
\usepackage[ruled,vlined]{algorithm2e}
\usepackage{nicefrac}       
\usepackage{microtype}      
\usepackage{amsthm}
\usepackage{amsfonts}
\usepackage{bm}
\usepackage{gensymb}
\usepackage{enumitem}
\usepackage{tikz}
\usepackage{multirow}
\usepackage{array}
\usepackage{makecell}
\usepackage{listings}
\usepackage{capt-of}
\usepackage{ragged2e}

\usepackage[noabbrev,nameinlink]{cleveref}
\crefname{property}{property}{Property}
\creflabelformat{property}{(#1)#2#3}
\crefname{equation}{eq}{Eq}
\creflabelformat{equation}{(#1)#2#3}

\makeatletter
\newcommand{\thickhline}{%
    \noalign {\ifnum 0=`}\fi \hrule height 1.3pt
    \futurelet \reserved@a \@xhline
}

\newtheorem{theorem}{Theorem}[section]
\newtheorem{corollary}{Corollary}[theorem]
\newtheorem{lemma}[theorem]{Lemma}

\newtheorem{fact}{Fact}
\theoremstyle{definition}
\newtheorem{definition}{Definition}[section]

\newcolumntype{x}[1]{>{\centering\arraybackslash\hspace{0pt}}p{#1}}

\title{On the Robustness and Generalization of Deep Learning Driven Full Waveform Inversion}

\author{
  Chengyuan Deng \\
  Los Alamos National Laboratory\\
  \texttt{charles.deng@lanl.gov} \\
  \And
  Youzuo Lin \\
  Los Alamos National Laboratory \\
  \texttt{ylin@lanl.gov} \\
}

\begin{document}
\maketitle

\begin{abstract}
  The data-driven approach has been demonstrated as a promising technique to solve complicated scientific problems. Full Waveform Inversion~(FWI) is commonly epitomized as an image-to-image translation task, which motivates the use of deep neural networks as an end-to-end solution. Despite being trained with synthetic data, the deep learning-driven FWI is expected to perform well when evaluated with sufficient real-world data. In this paper, we study such properties by asking: how robust are these deep neural networks and how do they generalize? For robustness, we prove the upper bounds of the deviation between the predictions from clean and noisy data. Moreover, we demonstrate an interplay between the noise level and the additional gain of loss. For generalization, we prove a norm-based generalization error upper bound via a stability-generalization framework. Experimental results on seismic FWI datasets corroborate with the theoretical results, shedding light on a better understanding of utilizing Deep Learning for complicated scientific applications.
\end{abstract}


\section{Introduction}
\label{sec:Intro}
\begin{figure}
    \centering
    \includegraphics[width=15cm]{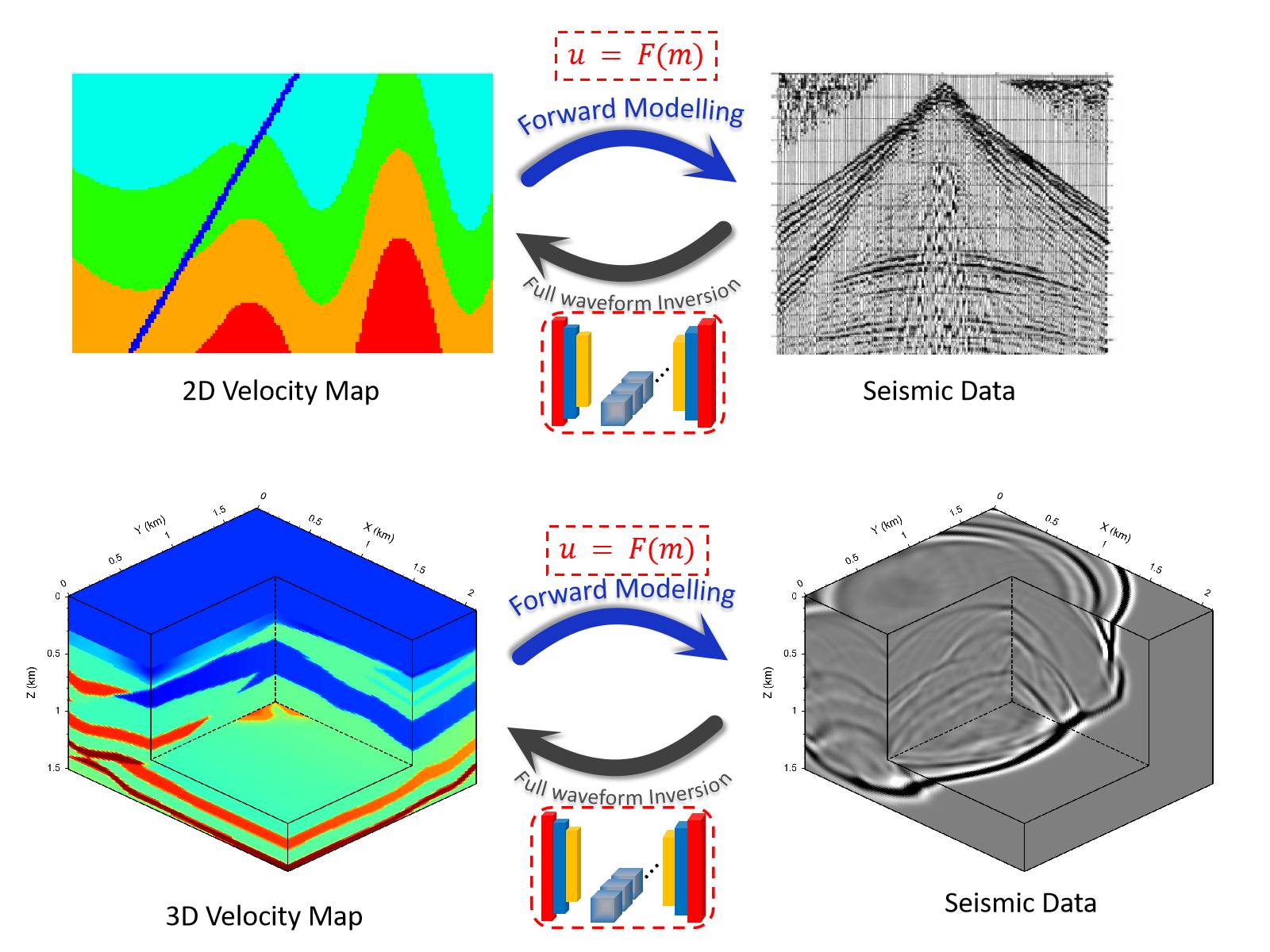}
    \caption{Seismic FWI (2D \& 3D) and Forward Modeling}
    \justifying
    \textnormal{\cref{fig:fwi} shows an example of seismic FWI with the relationship between velocity maps and seismic data. Forward modelling takes velocity maps as input and calculates the seismic signals with PDEs. While in practice, the seismic signals can be recorded by geological equipment, the FWI is expected to produce the velocity maps so that people can understand the subsurface geologic structures. } 
    \label{fig:fwi}
\end{figure}

The surge of interest in exploiting Deep Learning for difficult scientific problems has been witnessed with remarkable success in physics \cite{willard2020integrating,mehta2019high},  geoscience \cite{wu2019inversionnet, yang2019deep},  and neuroscience \cite{richards2019deep, zhu2019applications}, etc. Usually combined with physics-informed constraints, the data-driven approaches typically solve the problems via an end-to-end manner, as a circumvention of the expensive computation and ill-posedness issue inherited from those problems. Full-waveform Inversion (FWI), emerges as a classical inverse problem as described \cite{virieux2009overview, xu2012full, tarantola2005inverse}. The FWI explores geophysical properties such as site geology, stratigraphy, and rock quality employing reconstructing subsurface velocity models from seismic waveform signals. It aims at minimizing the misfit between the predicted and recorded seismic waveforms, thus lies in the family of optimization problems constrained by partial differential equations (PDEs). \Cref{fig:fwi} presents an example of the FWI and its corresponding forward modeling.\par

En route to mitigating the cycle-skipping and ill-posedness issues of FWI, researchers have revolutionized the methodology from \textit{``physics-driven''}, which usually uses gradient-based optimization and incorporates physics constraints as a regularization term\cite{lin2014acoustic, lin2015quantifying},  to \textit{``data-driven''} approaches\cite{wu2019inversionnet, zeng2021inversionnet3d, wang2018velocity, li2019deep}. The latter has demonstrated significant empirical superiority by utilizing deep neural networks. As demonstrated in \cref{fig:fwi}, FWI is modelled as an image-to-image translation task, therefore sharing similarity with a myriad of domains in Computer Vision:  style transfer\cite{gatys2015neural,gatys2016image, johnson2016perceptual}, image super-resolution~\cite{dong2015image,yang2010image}, image restoration~\cite{zhang2017learning, lehtinen2018noise2noise}, etc. Inspired by such observations, most current works introduce celebrated deep learning models such as Convolutional Neural Networks~\cite{wu2019inversionnet,yang2019deep}, Generative Adversarial Networks~\cite{zhang2019velocitygan, goodfellow2014generative}, etc. as the backbone, which learns the target velocity map in an end-to-end manner. This practice, however, evokes two concerns to be evaluated more discreetly: robustness and generalization behavior of the proposed deep learning models.\par

The issue of robustness and generalization of learning algorithms has been studied in theoretical machine learning regimes for decades~\cite{mcallester1999pac, vapnik2015uniform, bartlett1998sample, xu2012robustness}, followed by empirical techniques to ameliorate both properties~\cite{wang2020sag, bhagoji2018enhancing, yoshida2017spectral, miyato2018spectral}. In recent years, people have extended the analysis to deep neural networks~\cite{long2019generalization, neyshabur2017exploring, arora2018stronger, cao2019generalization}, shedding light on understanding the obstacles for deep learning to perform well in more general scenarios. For deep learning-driven FWI, such difficulties are exacerbated due to the following reasons. (1) Acquiring and processing real-world data requires arduous human efforts and an excessively long time~(sometimes hundreds of years), therefore most DNNs for FWI are learned and evaluated on synthetic datasets. (2) The recorded seismic data for testing is usually contaminated by noise, however, the seismic data is synthesized from clean signals. (3) The training data is often synthesized with fixed geophysical features, such as source signature, wave frequency, however, there might be a slight shift on those features during testing, thus raising the issue of generalization with potential distribution drift. Furthermore, as elaborated in \cref{sec:review}, most of the proposed DNN models \cite{wu2019inversionnet, wang2018velocity, yang2019deep, zeng2021inversionnet3d} are based on an encoder-decoder architecture, thus sharing a possibility to perform similarly. Because of the above evidence, the desiderata grows to understand the robustness and generalization of DNNs for FWI problems, thus motivating the following questions: \par

\begin{center}
    \textit{How robust are DNN models for FWI when tested with noisy data? \\
How well do these DNN models generalize?}
\end{center}

In this paper, we would like to answer the above questions. In the first question, the robustness of DNNs is defined as the performance degradation when they are tested with perturbed noisy data. We analyze deep neural networks trained with MAE loss and MSE loss, respectively. An upper bound of the loss gain is proved based on Lipschitz continuity and H\"{o}lder's inequality for MAE loss. The bound indicates that the performance degradation is associated with the product of weight matrices at each layer and the noise itself. A corollary implies that the bound of MSE loss is looser than that of MAE loss as the noise level raises.  The analysis of generalization is established upon the framework proposed by \cite{xu2012robustness}, which bridges a stability property of an algorithm to its generalization error. An upper bound of the generalization error is obtained, taking the norm of weights and the covering number as important parameters. \par
To provide intuitions on the theoretical results and connect them to practice, we conduct a meticulous empirical study on both topics by training an encoder-decoder deep neural network on standard seismic FWI datasets. In the robustness test, we illustrate the interplay between the perturbation intensity (SNR) and the prediction performance change for DNNs with MAE loss and MSE loss. The conclusion is that DNNs with MSE loss give worse predictions as the noise increases. The experiments on generalization are developed in two settings: (1) standard-setting, where the generalization gap is defined as the difference between train loss and test loss, provided the data samples are i.i.d in both datasets, and (2) FWI-domain-setting, where the test data has a slight distribution drift caused by instability of the geological features in real-world data. We show that our generalization bounds have a positive relationship with the empirical generalization gap in both settings.

We summarize the contributions of this paper as follows:

\begin{enumerate}
    \item Upper bounds of gain of loss for Deep Neural Networks tested with noisy data.
    \item An upper bound of generalization error for Deep Neural Networks solving FWI in an end-to-end manner.
    \item In addition to the context of FWI, our results and analysis can be extended to other inverse problems and computational imaging tasks with some adjustment, thus may potentially inspires broad interest in the inverse problems community.
\end{enumerate}
The rest of this paper is arranged as follows: In Section~\ref{sec:bg}, we introduce the necessary backgrounds and the notations; Section~\ref{sec:robust} focuses on robustness, and presents the upper bound of the difference when testing on noisy and clean data; Section~\ref{sec:generalization} moves forward to generalization, which is demonstrated with the generalization error bound; Section~\ref{sec:review} provides an overview of the research lines on deep learning-driven FWI and theoretical analysis of DNNs; In Section~\ref{sec:experiments}, we show empirical results implied by our theoretical results; Section~\ref{sec:conclusion} concludes the paper and proposes open problems.


\section{Related Work}
\label{sec:review}
\subsection{Deep Learning Driven FWI}
With the recent progress of deep learning in image generation tasks, researchers have adopted deep neural networks as the first-of-choice as the data-driven solution for FWI. GeoDNN~\cite{araya2018deep} proposed an 8-layer fully-connected neural network. Other attempts include InversionNet~\cite{wu2019inversionnet},  modifiedFCN~\cite{wang2018velocity} and FCNVMB~\cite{yang2019deep}, all of which are encoder-decoder networks with CNN as the backbone. SeisInvNet~\cite{li2019deep} enhanced each seismic trace with auxiliary knowledge from neighborhood traces for better spatial correspondence. NNFWI~\cite{zhu2020integrating}  used deep models to generate a physical velocity model, which is then fed to a PDE solver to simulate seismic waveforms. All the mentioned work focused on 2D modeling, and we refer to a through survey~\cite{adler2021deep} for complete references. Recently, Zeng et al. proposed InversionNet3D~\cite{zeng2021inversionnet3d} based on an encoder-decoder network, as the first deep learning solution for high-resolution 3D FWI. InversionNet3D employs group convolution in the encoder and invertible layers in the decoder to achieve high efficiency and scalability. A fewer other works based on recurrent neural networks(RNNs)~\cite{richardson2018seismic} are proposed as a contrast to CNNs. Unsupervised learning techniques are also exploited for representation learning and data augmentation for FWI~\cite{Jin-2021-Unsupervised, yang2021making, zhang2019velocitygan,marcus2021data}.

\subsection{Generalization of Deep Neural Networks}
There has been substantial progress on characterizing the generalization behavior of machine learning algorithms, both empirically and theoretically. The earliest works were propelled from a theoretical perspective of understanding the uniform convergence of empirical quantities to their mean. Several complexity measure frameworks were established then, and maintain as essential ingredients even in modern analysis: VC dimension~\cite{vapnik1974theory, vapnik1991necessary, evgeniou2000regularization}; Rademacher complexity~\cite{bartlett2002rademacher, bartlett2005local}, fat-shattering dimension~\cite{bartlett1998sample, alon1997scale}, and PAC-Bayes bound~\cite{mcallester1999pac}. Recently, initiated by \cite{neyshabur2015norm}, a family of norm-based bounds has emerged~\cite{neyshabur2015norm, bartlett2017spectrally, neyshabur2017pac, li2018tighter, golowich2018size, lin2019generalization}. This family employs the product of Frobenius norm or spectral norm of each layer as a crucial factor of the generalization bound. Notice that the norm-based bounds can be generally applied to both fully-connected neural networks~(FNNs) and convolutional neural networks~(CNNs). \cref{tab:geb_norm} demonstrates a comparison of the norm-based bounds for both FNNs and CNNs. There are also independent works from various views. Long et al.\cite{long2019generalization} proved a bound for deep CNNs related to the total number of parameters and the distance from trained weights to initial weights. Ledent et al.\cite{ledent2021normbased} incorporated weight-sharing in CNNs and proved a bound for multi-classification setting.

\begin{table}[]
    \centering
    \begin{tabular}{|c|l|l|}
        \specialrule{.1em}{.05em}{.05em} 
          Work & Simplified Bounds & Remark  \\ 
          \specialrule{.1em}{.05em}{.05em} 
         \cite{neyshabur2015norm} & $\Tilde{O}\bigg(2^d\prod_{i=1}^{d}||W_i||_{F}/\sqrt{n}\bigg)$ & CNN: $W_i \rightarrow op_i$\\
         \hline
         \cite{bartlett2017spectrally} & $\Tilde{O}\bigg( \prod_{i=1}^{d}||W_i||_\sigma \big( \sum_{i=1}^{d}\frac{||W_i||_{2,1}^{2/3}}{||W_i||_\sigma^{2/3}}\big)^{3/2}/\sqrt{n}\bigg)$ & CNN: $W_i \rightarrow op_i$\\
         \hline
         \cite{neyshabur2017pac} & $\Tilde{O}\bigg( \prod_{i=1}^{d}||W_i||_\sigma \sqrt{d^2w\sum_{i=1}^{d}\frac{||W_i||_F^{2}}{||W_i||_{\sigma}^{2}}}/\sqrt{n}\bigg)$  & CNN: $W_i \rightarrow op_i$, $w \rightarrow cm$\\
         \hline
         \cite{golowich2018size} & $\Tilde{O}\bigg( \prod_{i=1}^{d}||W_i||_F \cdot \min \{ 1/\sqrt[4]{n}, \sqrt{d/n}\}\bigg)$ & CNN: $W_i \rightarrow op_i$\\
         \hline
         \cite{li2018tighter}  &  $\Tilde{O}\bigg( \prod_{i=1}^{d}||W_i||_\sigma \sqrt{d^2w}/\sqrt{n}\bigg)$  & $d\gg w$, CNN: $W_i \rightarrow op_i$, $w \rightarrow cm$ \\ 
         \hline
         \cite{lin2019generalization} & $\Tilde{O}\bigg( \prod_{i=1}^{d}||W_i||_{\sigma}^{1/4}\big(d^2w^4 \sum_{i=1}^{d}\frac{||W_i||_F^{2}}{||W_i||_{\sigma}^{2}}\big)^{1/4}\bigg)$ & CNN: $W_i \rightarrow op_i$, $w^4 \rightarrow r^2c^2\sqrt{m}$\\ 
         \specialrule{.1em}{.05em}{.05em} 
    \end{tabular}
    \caption{A Comparison of Norm-based bounds}
    \justifying
    \textnormal{Notations: $n$ is the number of training samples, $d$ is the number of layers, $W_i$ is either the weight matrix for FNNs or the linear operator $(op_{i})$ for CNNs of the \textit{i-th} layer and $w$ is the largest layer width; $||\cdot||_F$ denotes the Frobenius norm and $||\cdot||_\sigma$ denotes the spectral norm. For CNNs, $c$ denotes number of channels, $r$ represents the size of filter, $m$ is the number of outputs generated by the network.}
    \label{tab:geb_norm}
\end{table}

Another perspective of understanding the generalization behavior of DNNs is empirical study. Arora et al.\cite{arora2018stronger} observed the noise-stability of weight matrices as the networks go deeper via massive experiments, and proposed generalization bounds via a compression framework. Also by proposing the remarkable double descent phenomena, Nakirran et al.\cite{nakkiran2019deep} suggested that larger model and size of training data may not be helpful. Notably, Jiang et al. \cite{jiang2019fantastic} did a thorough empirical study on three most representative types of generalization error bounds.

\section{Backgrounds}
\label{sec:bg}
We firstly introduce definitions of the essential ingredients and formalizing the deep learning models in the regime of FWI and, along the way, bringing in the notations. Throughout this paper, we denote $\mathcal{X} \subseteq \mathbb{R}^{n_x}$ and $\mathcal{Y} \subseteq \mathbb{R}^{n_y}$ as the real spaces, the inverse problem is to estimate $x\in \mathcal{X}$ from $y\in \mathcal{Y}$ with $\mathcal{F}^{-1}$ such that:
\begin{displaymath}
y = \mathcal{F}(x)+n,
\end{displaymath}
where $n$ is usually assumed as Gaussian noise.

\subsection{Full Waveform Inversion}
The Full Waveform Inversion (FWI) targets the optimal velocity map $m$ to minimize the difference between predicted and observed seismic data $u = \mathrm{u}(x, t)$, where $x$ is the location and $t$ refers to a timestamp. Typically, the acoustic FWI follows the following PDE constriant:

\begin{align} \label{equ:pde}
    \frac{\partial^2 u}{\partial t^2} = \nabla \cdot (m^2 \nabla) +f,
\end{align}
where $f$ is the source term. 

Focusing on the partial differential equation, the forward modeling can be formulated in the form of: $\mathcal{F}(m) = u$, naturally the inverse problem is given by $m = \mathcal{F}^{-1}(u)$. The optimization task for such an inverse problem can be expressed as:
\begin{align}
    \min_m \mathcal{L}(\hat{u}(x,t,m), u) \textit{ s.t. } F_{acoustic}(u,m)=0.
\end{align}
where $\mathcal{L}(\cdot)$ denotes the loss function, $\hat{u}$ is the estimated seismic data, and $F_{acoustic}(u,m)$ stands for the PDE in~\cref{equ:pde}

\subsection{Deep Neural Networks}
We consider three variants of DNNs: fully-connected neural networks (FNNs), general convolutional neural network (CNNs), and encoder-decoder convolutional neural networks (usually with all convolutional layers), as they are the first-of-choice in FWI problem and other computational imaging tasks~\cite{wu2019inversionnet,yang2019deep, li2019deep, wang2018velocity}. To simplify the analysis, we will show in this section that the three variants share a unified formulation. \par

For a neural network $G$ with $d$ layers, denote ($\sigma_1, \dots, \sigma_d$) as the activation function (e.g. ReLU) or pooling function after each layer, with the assumption that $\sigma_i$ is $l_i$-Lipschitz continuous.
Recall that an inverse problem takes $y\in \mathcal{Y}$ as the input, for a fully-connected neural network, we obtain:
\begin{equation} \label{equ:fnn}
G_\theta(y) := \sigma_d\Big( W_d\sigma_{d-1}\big(W_{d-1} \dots \sigma_1(W_1\cdot y + b_1)\dots \big)   +b_d\Big),
\end{equation}
where $\theta = \{ (W_i, b_i): \big| W_i \in \mathbb{R}^{n_i \times n_{i-1}}, b\in \mathbb{R}\}$.

It has been demonstrated in \cite{lin2019generalization, sedghi2019singular} that convolution is a linear operation, thus can be represented as matrix multiplication. Formally, we have the following:

\begin{fact} \label{fact:conv}
Suppose a convolutional filter with dimension $d_c$ is imposed on input $x\in \mathbb{R}^{d_{in}}$, let $W \in \mathbb{R}^{1 \times d_c}$ be the weight matrix, then there exists a unique matrix $\mathrm{op} \in \mathbb{R}^{d_{in} \times d_{out}}$ such that $conv(x, W) = \mathrm{op} \cdot x$ and $\forall \mathrm{op}_{i,j}$ is either 0 or $w_k \in W$.
\end{fact}

In an encoder-decoder-based CNN, the down-sampling process is accomplished by a stack of convolution layers, while the up-sampling hinges on the decoder, which is generally composed of transposed convolution layers~\cite{dumoulin2016guide}. It is worth mentioning that there are other up-sampling approaches such as fractional convolution~\cite{radford2015unsupervised, shi2016deconvolution} or backward convolution~\cite{dosovitskiy2015flownet, long2015fully}. A thorough discussion in \cite{shi2016deconvolution} shows that they generate identical results if the filter is learned. Thus, we focus on the transposed convolution commonly used. The following fact states that the transposed convolution operator also applies a linear transformation to the input. 
\begin{fact} \label{fact:deconv}
Following \cref{fact:conv}, given $x, W$ and output $y\in \mathbb{R}^{d_{out}}$ such that $y = conv(x, W) = \mathrm{op} \cdot x$, then there exists another unique matrix $\mathrm{op}'$ such that $x = deconv(y, W) = \mathrm{op}' \cdot x$
\end{fact}

\cref{fact:deconv} is demonstrated in \cite{shi2016deconvolution} with under context of image super resolution. For both \cref{fact:conv} and \cref{fact:deconv}, we provide an example in the appendix.\par

Because of the above facts, we denote $\mathrm{op}_i$ as the matrix generated by the convolution operator imposed on $X_{i-1}$, so that Eq.~\cref{equ:fnn} can be applied to encoder-decoder convolutional neural networks by setting $b_i=0$. Specifically, denote ($W_1, \dots W_d$) as the weights of convolutional kernels at each layer, $W_i \in \mathbb{R}^{c_i \times r_i}$ has $c_i$ convolutional kernels, each with kernel size $r_i$. Letting $y_i$ be the output of layer $i$, the following holds:
\begin{displaymath}
    y_{i+1} = \sigma_{i+1}\big( \mathrm{op}_{i+1}(W_{i+1}y_i)\big).
\end{displaymath}
Therefore, the class of encoder-decoder convolutional neural networks follows:
\begin{equation}
    G_\theta(y) := \sigma_d\Big( \mathrm{op}_{d}W_d\sigma_{d-1}\big(\mathrm{op}_{d-1}W_{d-1} \dots \sigma_1(\mathrm{op}_{1}W_1\cdot y)\dots \big)   \Big).
\end{equation}

The family of general convolutional neural networks is mostly a combination of fully-connected layers and convolutional layers, together with activation and pooling functions. Hence we use $(W_1, \dots, W_d)$ as either the weight matrices of layers in fully-connected neural networks, or the convolution operator with weights of convolutional layers, and obtain a unified formulation of DNNs in the regime of FWI:
\begin{equation}
    G_\theta(y) := \sigma_d\Big( W_d\sigma_{d-1}\big(W_{d-1} \dots \sigma_1(W_1\cdot y)\dots \big)   \Big).
\end{equation}

\subsection{Definitions}
\begin{definition}
(\textbf{Lipschitz Continuity}) A real-valued function $f: \mathbb{R}^m \longrightarrow \mathbb{R}^n$ is $L-lipschitz$ if there exists a real constant $L>0$, s.t. $\forall x_1, x_2 \in \mathbb{R}^m$, 
\begin{align*}
    |f(x_1)-f(x_2)| \leq L|x_1-x_2|.
\end{align*}
\end{definition}

\begin{definition} \label{def:coverno}
(\textbf{Covering number}) Let $(X, \rho)$ be a metric space and $H \subseteq X$, we say that $G$ is $\epsilon -cover$ of $H$ with respect to $\rho$ if every $h\in H$ has a $g\in G$ such that $\rho(g,h) \leq \epsilon$. Then $\mathcal{N}_\rho (H, \epsilon)$ denotes the size of smallest $\epsilon -cover$ of $H$ \textit{w.r.t.} $\rho$.  
\end{definition}


\begin{definition}
(\textbf{Jacobian matrix}) Consider a $d$-layer fully-connected neural network parameterized by $\theta$ with $1$-lipshitz activation functions $\phi(\cdot)$: $\hat{x} = \mathbf{G}_\theta (y) = W_d^T \phi(\dots \phi (W_1^T y+b_1) \dots) +b_d $, the Jacobian matrix of the neural network $\mathbb{G}_\theta$ is given as:
\begin{equation*}
    J = \frac{d\hat{x}}{dy}=
    \begin{bmatrix}
    \frac{\partial \hat{x_1}}{y_1} & \dots & \frac{\partial \hat{x_1}}{y_{N_y}} \\
    \vdots & \ddots & \vdots \\
    \frac{\partial \hat{x}_{N_x}}{\partial y_1} & \dots & \frac{\partial \hat{x}_{N_x}}{\partial y_{N_y}}
    \end{bmatrix}.
\end{equation*}
\end{definition}
Owing to the chain rule of derivatives, $J$ can be decomposed into the product of layer-wise Jacobian matrices \textit{i.e.} $J=\prod_{i=1}^d J_i$.

\begin{definition} \label{def:robust}
($\mathbf{K}-\epsilon (\mathbf{T})$ \textbf{robustness}) Let $\mathbf{T}$ be the training set of $N$ entries and $\mathbf{D}$ be the sample space. A learning algorithm $\mathbf{G}$ is said to be $\mathbf{K}-\epsilon (\mathbf{T})$ robustness if $\mathbf{D}$ can be partitioned into $\mathbf{K}$ disjoint sets $\mathcal{K}_k, (k = 1, \dots K) $, such that for any $(x_i, y_i) \in \mathbf{T}$ and $(x, y) \in \mathbf{D}$:
\begin{equation}
    (x_i, y_i), (x,y) \in \mathcal{K}_k \Longrightarrow |\mathcal{L}(G(y_i), x_i)-\mathcal{L}(G(y), x)| \leq \epsilon(\mathbf{T}).
\end{equation}
\end{definition}


\section{Robustness Against Perturbation}
\label{sec:robust}
In this section, we study how the performance of deep neural networks might be impacted when tested with noisy data. The gravity of this issue increases for FWI as the test data collected from the real world is inevitably contaminated with noise. We consider DNNs trained with MAE loss and MSE loss and analyze how the model robustness differs from each other. By the end of this section, we conclude that as the noise level increases, DNNs trained with MAE loss show better robustness than DNNs trained with MSE loss.
\subsection{Setup}
 Throughout the analysis, we make merely one assumption about the level of noise is bounded. In other words, the random noise $n$ satisfies: $||n||_2 \leq \eta$. \par
Now let's consider a DNN $G_\theta$ which learns a mapping: $\mathbb{R}^p \longrightarrow \mathbb{R}^q$, its robustness indicates how significant the performance drop will be if the test data is imposed with random noise $n$. Denoting $\mathcal{L}(\cdot): \mathbb{R}^q \longrightarrow \mathbb{R}$ as the loss function adopted, we are interested in the upper bound of $|\mathcal{L}\big(G_\theta(x+n)\big)-\mathcal{L}\big(G_\theta(x)\big)|$.

\subsection{Upper Bound of DNNs Robustness with MAE}
As an essential ingredient of the optimization process, the loss functions commonly adopted for image-to-image analysis include Mean Absolute Error (MAE) or (Rooted) Mean Square Loss (MSE/RMSE)~\cite{qi2020mean}, with the exponential term, MSE implicitly gives a higher weight on large errors, thus becoming more sensitive to marginal predictions caused by perturbations on the input data. Inspired by such an inductive bias, we first consider MAE loss as the objective function of the DNNs and prove that the gain of the loss could be upper-bounded, thus indicating certain level of robustness can be guaranteed. The following theorem states our main result:
\begin{theorem} \label{thm:robust}
\textbf{(Robustness)} For a objective function $g = f \circ \mathcal{L}: \mathbb{R}^p \longrightarrow \mathbb{R}$, where $f$ denotes a neural network with $d$ layers that learns a mapping $f: \mathbb{R}^p \longrightarrow \mathbb{R}^q$ and $\mathcal{L}: \mathbb{R}^q \longrightarrow \mathbb{R}$ is loss function. If the input $x \in \mathbb{R}^p$ is perturbed data with an arbitrary noise $n\,(||n||_2 \leq \eta)$, then the following holds if $\mathcal{L}$ is MAE loss:
\begin{align*}
    RB_{MAE} = |g(x+n)-g(x)| \leq \prod_{i=1}^{d}||\mathbf{W}_i||_F \cdot \eta,
\end{align*}
\end{theorem}
where $\mathbf{W}_i$ is the weight matrix of the \textit{i-th} layer in the DNNs as defined in \cref{sec:bg}.\par
Before proving theorem \ref{thm:robust}, we introduce the following lemma:
\begin{lemma}
\label{lem:rob}
Consider a function $f: \mathbb{R}^d \longrightarrow \mathbb{R}$ that satisfies the Lipschitz continuity, we have:
\begin{align*}
    |f(x)-f(y)| \leq L_p ||x-y||_q,
\end{align*}
where $\frac{1}{p}+\frac{1}{q}=1, (1 < p,q < \infty)$, $L_p=\sup \left\{ ||\nabla f(x)||_p: x\in \mathbb{R}^d \right\}$.
\end{lemma}
\begin{proof}
\begin{align*}
    |f(x)-f(y)| &= |f(x_1, x_2, \dots, x_n)-f(y_1, y_2, \dots, y_n)|, \\
    &= |f(x_1, x_2,\dots, x_n)-f(y_1, x_2, \dots, x_n)+f(y_1, x_2, \dots, x_n)-f(y_1, y_2, \dots, y_n)|,\\
    &\leq |f(x_1, x_2,\dots, x_n)-f(y_1, x_2, \dots, x_n)| + |f(y_1, x_2,  \dots, x_n)-f(y_1, y_2, \dots, y_n)|,\\
    &= \frac{|f(x_1, x_2,\dots, x_n)-f(y_1, x_2, \dots, x_n)|}{|x_1-y_1|}\cdot |x_1-y_1|+\\
     & |f(y_1, x_2,  \dots, x_n)-f(y_1, y_2, \dots, y_n)|.
\end{align*}
From the middle values theorem, there exists a constant $c_1$ such that the first term equals: $|f^{'}_{x_1}(c_1)|\cdot |x_1-y_1|$. Now we focus on the second term, with triangular inequality and middle values theorem again: 
\begin{align*}
    &|f(y_1, x_2,  \dots, x_n)-f(y_1, y_2, \dots, y_n)| \\
    &\leq |f(y_1, x_2,  \dots, x_n)-f(y_1, y_2, x_3, \dots, x_n)|+|f(y_1, y_2, x_3, \dots, x_n)-f(y_1, y_2, \dots, y_n)|,\\
    &= \frac{|f(y_1, x_2,\dots, x_n)-f(y_1, y_2, \dots, x_n)|}{|x_2-y_2|}\cdot |x_2-y_2|+|f(y_1, y_2, x_3, \dots, x_n)-f(y_1, y_2, \dots, y_n)|,\\
    &= |f^{'}_{x_2}(c_2)|\cdot |x_2-y_2|+|f(y_1, y_2, x_3, \dots, x_n)-f(y_1, y_2, \dots, y_n)|.
\end{align*}

So on and so forth, by substituting $x_i$ with $y_i$ on the second term iteratively, and summing all the terms by middle values theorem, we obtain:
\begin{equation}
\label{equ:mv}
    |f(x)-f(y)| \leq \sum_{i=1}^{n}|f^{'}_{x_i}(c_i)|\cdot |x_i-y_i|.
\end{equation}
Now recall the H\"{o}lder's inequality:
\begin{fact}
(H\"{o}lder's Inequality)
\begin{align*}
    \sum_{i=1}^{n} |a_ib_i| \leq (\sum_{i=1}^{n}|a_i|^p)^\frac{1}{p} \cdot (\sum_{i=1}^{n}|b_i|^q)^\frac{1}{q},
\end{align*}
where $p,q >1$ and $\frac{1}{p}+\frac{1}{q}=1$, equality holds when $|b_i|=c|a_i|^{p-1}$.
\end{fact}
Equation \ref{equ:mv} can be extended as:
\begin{align*}
    |f(x)-f(y)| &\leq \sum_{i=1}^{n}|f^{'}_{x_i}(c_i)|\cdot |x_i-y_i|, \\
    &\leq (\sum_{i=1}^{n}|f^{'}_{x_i}(c_i)|^p)^{\frac{1}{p}} \cdot (\sum_{i=1}^{n}|x_i-y_i|^q)^\frac{1}{q},\\
    &= ||\nabla f(c)||_p \cdot ||x-y||_q = L_p \cdot ||x-y||_q,
\end{align*}
where $c = (c_1, c_2, \dots, c_n) \in \mathbb{R}^d$.
\end{proof}

Lemma \ref{lem:rob} requires that $f$ must be Lipschitz-continuous, which characterized the predictions of our deep neural networks are bounded  with respect to the input. Recent progress~\cite{fazlyab2019efficient, scaman2018lipschitz} has shown that DNNs with Relu activations are indeed Lipchitz-continuous, we employ this evidence as the last piece to complete the puzzle, 
and now we are ready to present the proof of theorem \ref{thm:robust}.
\begin{proof}
Recall that the loss function $\mathcal{L}$ is specified as MAE loss, which indicates $\mathcal{L}(\cdot) = ||\cdot||_1$. 
\begin{align*}
    |g(x+n)-g(x)| &= |||f(x+n)||_1 - ||f(x)||_1| \\
    &\leq |f(x+n)-f(x)|, \tag*{(triangular inequality)} \\
    &\leq L_2 \cdot ||x-y||_2 \tag*{(lemma \ref{lem:rob})},\\
    &\leq \prod_{i=1}^{d}\mathbf{|J_i|}\cdot \eta,\\
     &\leq \prod_{i=1}^{d}\mathbf{||W_i||_F}\cdot \eta,
\end{align*}
which completes the proof.
\end{proof}

An immediate observation on \cref{thm:robust} is that the performance degradation is bounded by a constant. If a pre-trained model is tested with noisy data, $|J_i|$ or $||W_i||_F$ is fixed and the performance depends entirely on the noise level. While as argued in \cite{zeng2021inversionnet3d}, the additive noise is considered as Gaussian noise at a low level in most scenarios, thus leading to a significant performance degradation with low probability. \par

\subsection{Upper Bound of DNNs Robustness with MSE}
It can be observed that the technique above is not applicable to the proof of MSE loss. Let $\mathcal{L}(x, \hat x)$ denote the loss of the predicted values and the ground truth, it is trivial that $|\mathcal{L}(x_1, \hat x)-\mathcal{L}(x_2, \hat x)| \leq \mathcal{L}(x_1, x_2)$ when $\mathcal{L}$ stands for MAE loss, namely $||\cdot||_1$. Equipped with such a fact, we can apply lemma \ref{lem:rob}. However this requirement cannot be satisfied by MSE loss. On the contrary, we can show that  $|\mathcal{L}(x_1, \hat x)-\mathcal{L}(x_2, \hat x)| > \mathcal{L}(x_1, x_2)$ for certain values of $x$. We formalize the fact for MSE loss as \cref{fact:mse} and provide a proof in the appendix.\par

\begin{fact} \label{fact:mse}
For $\forall x_1, x_2 \in \mathbb{R}^d$, and $\mathcal{L}(x, \hat x) = ||x-\hat x||_2$, $|\mathcal{L}(x_1, \hat x)-\mathcal{L}(x_2, \hat x)| \leq \mathcal{L}(x_1, x_2)$ does not hold for $\forall x \in \mathbb{R}^d$
\end{fact}

However, there is a simple approach to upper bound the gain of MSE loss based on \cref{thm:robust}, whilst implying that $RB_{MSE}$ is looser than $RB_{MAE}$. An ingredient here is that a neural network is lipschitz continuous with any common activation function (ReLU, tanh, sigmoid, etc.)~\cite{fazlyab2019efficient, scaman2018lipschitz}. The bound is given in the following \cref{cor:mse}
\begin{corollary} \label{cor:mse}
For a objective function $g = f \circ \mathcal{L}: \mathbb{R}^p \longrightarrow \mathbb{R}$, where $f$ denotes a neural network with $d$ layers that learns a mapping $f: \mathbb{R}^p \longrightarrow \mathbb{R}^q$ and $\mathcal{L}: \mathbb{R}^q \longrightarrow \mathbb{R}$ is loss function. If the input $x \in \mathbb{R}^p$ is perturbed with an arbitrary noise $n\,(||n||_2 \leq \eta)$, then the following holds if $\mathcal{L}$ is MSE loss:
\begin{displaymath}
RB_{MSE} \leq L \cdot \frac{\eta}{\sqrt{d_{in}}} \cdot (RB_{MAE}+2a),
\end{displaymath}
where $L$ is the Lipschitz constant of the neural network, $d_{in}$ is the dimension of input data, and $a$ denotes that maximum of test loss on clean data, all of which are considered as a constant.
\end{corollary}
\begin{proof}
\begin{align*}
    RB_{MSE} & = |g(x+n)-g(x)|, \\
    & = \Big|||f(x+n)-y||_2 - ||f(x)-y||_2\Big|, \\
    & \leq |f(x+n)+f(x)|\cdot |f(x+n)-f(x)| - 2y\cdot |f(x+n)-f(x)|, \\
    & = |f(x+n)-f(x)|\cdot |f(x+n)+f(x)-2y|.
\end{align*}
Suppose $L$ is the Lipschitz constant of the neural network, following the definition we have: $|f(x+n)-f(x)|\leq L\cdot |n|$, let $a$ be the max test loss with clean data, namely $\max \big({|f(x)-y|}\big)$
\begin{align*}
    RB_{MSE} &\leq L\cdot |n| \cdot \big(|f(x+n)-y|+|f(x)-y|\big), \\
    &\leq  L\cdot |n| \cdot (RB_{MSE}+2|f(x)-y|),\\
    &\leq  L\cdot \frac{\eta}{\sqrt{d_{in}}}\cdot (RB_{MSE}+2a), \tag*{(Cautchy-schwartz Inequality)}
\end{align*}
concluding the proof.
\end{proof}

We would like to remark that the Lipschitz constant $L$, although intractable, has been shown to yield a close estimation for both Fully-connected Neural networks~\cite{scaman2018lipschitz, fazlyab2019efficient, szegedy2013intriguing, combettes2020lipschitz} and Convolutional Neural Networks~\cite{zou2019lipschitz, balan2018lipschitz, bartlett2017spectrally}. Notably, \cite{fazlyab2019efficient} proposed an estimation $L \approx \sqrt{\rho}$ where $\rho$ can be obtained via semi-definite programming. \par

\subsection{DNNs with MAE Are More Robust with Significant Noise}
Notice that although the bound of $RB_{MAE}$ is tighter than that of $RB_{MSE}$, we are still not close to the conclusion that DNNs with MAE is more robust than DNNs with MSE. First of all, the bounds are deterministic, so that no claim on an instance can be made. This is trivial if we consider an example that given $x\leq 100,\,y\leq 1000$, we may not conclude $x\leq y$. Secondly, our proof of \cref{thm:gen} and \cref{cor:mse} is the worst-case analysis, therefore obstructing probabilistic approaches. \par

To address this issue, we adopt the strategy of lower-bounding, $RB_{MSE}$. The insight is that it can only be concluded a bound induced by $RB_{MAE}$ is tighter than a bound induced by $RB_{MSE}$ if $LB(RB_{MSE}) \geq UB(RB_{MAE})$ where $UB$ stands for upper bound and $LB$ for lower bound. In the following corollary we show that $LB(RB_{MSE}) \geq UB(RB_{MAE})$ holds under certain conditions.
\begin{corollary}\label{cor:cond}
 $RB_{MSE} \geq RB_{MAE}$, if $|f(x+n)+f(x)-2y|\geq 1$.
\end{corollary}
\begin{proof}
For $RB_{MAE}$, we have $|g(x+n)-g(x)|\leq |f(x+n)-f(x)|$ from the proof of \cref{thm:robust};  For $RB_{MSE}$, let $g'$ denote the function of $f\circ \mathcal{L}$ when $\mathcal{L} = ||\cdot||_2$, then $|g'(x+n)-g'(x)| \geq |g'(x+n)|-|g'(x)|$ = $(f(x+n)+f(x)-2y)\cdot (f(x+n)-f(x))$; Now it is clear to see that $|f(x+n)+f(x)-2y|\geq 1$ must hold for $RB_{MSE}\geq RB_{MAE}$.
\end{proof}
From \cref{cor:cond}, it is straightforward that $|f(x+n)-y| + |f(x)-y| \geq 1$ must hold for $RB_{MSE}\geq RB_{MAE}$. There can be two scenarios: (1) The learning process has not converged, hence the predicted results will impose considerably large value of loss, namely both $|f(x+n)-y|$ and $|f(x)-y|$ can be large; (2) The training has already converged, then the test loss on clean data $|f(x)-y|$ is relatively small, as the additive noise $n$ increases, $|f(x+n)-y|$ also becomes incremental. Apparently, scenario (2) fits our context. Therefore, we are able to make a rigorous claim that DNNs with MSE become less robust than DNNs with MAE as the level of additive noise becomes more significant.


\section{Generalization Error Bound}
\label{sec:generalization}
Our analysis of generalization is established on the \textit{robustness and generalization} framework \cite{xu2012robustness}. Notice that the concept of robustness here deviates from  \cref{sec:robust}, as the robustness in \cite{xu2012robustness} is a property of learning algorithms applied to the entire data space and ours considers the testing phase. \par

\subsection{Setup}
In the context of supervised deep learning, training data $S_{train} = \{(x,y) | x \sim\mathcal{X}, y\sim\mathcal{Y}\}$ contains a number of $N$ \textit{i.i.d} samples from a sample space $S = X \times Y$ where $X \subset \mathbb{R}^p$ and $Y\subset \mathbb{R}^q$. Hence, in the training phase, an average training error can be obtained as $\mathbb{E}_{(x,y)\sim S_{train}}[\hat{\mathcal{L}}({G_\theta})]$ where $\hat{\mathcal{L}}(\cdot) = \frac{1}{N}\sum_{i=1}^{N}l(G_\theta(y_i), x_i)$. Similarly, by denoting the test error as $\mathbb{E}_{(x,y)\sim S}[\mathcal{L}({G_\theta})]$, the generalization error can be formally defined as $Err_g = |\mathbb{E}_{(x,y)\sim S_{train}}[\hat{\mathcal{L}}({G_\theta})]-\mathbb{E}_{(x,y)\sim S}[\mathcal{L}({G_\theta})]|$. Next, we will provide and further demonstrate an upper bound on $Err_g$.

\subsection{Main Result and Proof}
To Begin with, we present the main result by \cref{thm:gen}
\begin{theorem} \label{thm:gen}
\textbf{(Generalization)} For spaces $\mathbf{X}$ and $\mathbf{Y}$ equipped with $l_2$ norm, and sample space $\mathbf{D} = \mathbf{X} \times \mathbf{Y}$, with the assumption that the forward operator $\mathcal{F}()$ is $L$\textit{-Lipschitz}, a $d$-layer DNN. $\mathbf{G}_\theta (\cdot)$ learns a mapping: $\mathbf{Y} \longrightarrow \mathbf{X}$ trained on the training set $\mathbf{T}$ containing $N$ \textit{i.i.d} samples. Then with probability $1-\varepsilon$, the generalization error of $\mathbf{G_\theta}$ is upper-bounded by:
\begin{equation}
    Err_{G_{\theta}}\leq (1+\prod_{i=1}^d||W_i||_F)(L\delta + 2\eta)+M\sqrt{\frac{2\mathcal{N}(\frac{\delta}{2};\mathbf{X},l_2)\ln 2 + \ln (\frac{1}{\varepsilon})}{N}},
\end{equation}
where $L$ is the Lipschitz constant of the forward modelling, $\delta, \varepsilon, \eta$ are small constants, $M$ is also a constant representing the maximum training loss.
\end{theorem}
Recall \cref{def:coverno} and \cref{def:robust}, our first step is to prove that DNNs satisfy the $K-\epsilon (T)$ robustness with respect to the covering number of the data sample space as follows:

\begin{theorem} \label{thm:kepsilon}
Denote $(S, \rho)$ as a metric space where $\mathrm{S} = X \times Y$ is the sample space of the dataset, where $X \subset \mathbb{R}^p$ and $Y \subset \mathbb{R}^q$. Suppose a deep neural network which learns a mapping $f: Y \longrightarrow X$ is trained on a subset $S_{train} \subset S$, it follows that
\begin{displaymath}
\textnormal{the deep neural network is} \left(\mathcal{N}(\frac{\delta}{2};S,\rho), (1+\prod_{i=1}^{d}||J_i||_{2,2})\delta \right) \textnormal{-robust},
\end{displaymath}
where $||\cdot||_{2,2}$ denotes spectral norm, and $\mathcal{N}(\frac{\delta}{2};S,\rho)$ denotes the covering number of the metric space $(S, \rho)$ with $\frac{\delta}{2}$ as the radius of the metric ball.
\end{theorem}

\begin{proof}
The proof of \cref{thm:kepsilon} is an immediate result of the lemma below.

\begin{lemma} \label{lemma:spectral}
Suppose a deep neural network learns a inverse mapping $f: Y \longrightarrow X$, and $y_1, y_2 \in Y$, the following holds:
\begin{displaymath}
||G_\theta(y_1)-G_\theta(y_2)||_2 \leq \prod_{i=1}^d||J_i||_{2,2}\,||y_1-y_2||_2.
\end{displaymath}
\end{lemma}

To keep the consistency, we provide the proof of \cref{lemma:spectral} in \cref{apdx:mse}. We continue with the proof of \cref{thm:kepsilon} provided that \cref{lemma:spectral} is correct. 

Supposing that $s_1 = (x_1, y_1), s_2 = (x_2,y_2)$ are two samples from $S$ and $s_1 \in S_{train}$, we notice that it suffices to prove the following: 
\begin{displaymath}
|\mathcal{L}(G_\theta(y_1),x_1)-\mathcal{L}(G_\theta(y_2),x_2)| \leq (1+\prod_{i=1}^d||J_i||_{2,2})\cdot \delta,
\end{displaymath}
for a $\frac{\delta}{2}$-cover of $S$ and $\rho(s_1, s_2) \leq \delta$. \par

We adopt $l_2$ metric, and it follows that:
\begin{align*}
    |l(G_\theta(y_1),x_1)-l(G_\theta(y_2),x_2)|&=\Big| ||x_1-G_\theta(y_1)||_2-||x_2-G_\theta(y_2)_2 \Big|, \\
    & \leq ||x_1 - G_\theta(y_1) -x_2 + G_\theta(y_2)||_2, \tag*{(triangular inequality)} \\
    & \leq ||x_1-x_2||_2+||G_\theta(y_1)-G_\theta(y_2)||_2, \tag*{(Minkowski inequality)} \\
    & \leq ||x_1-x_2||_2+\prod_{i=1}^d||J_i||_{2,2}||y_1-y_2||_2, \tag*{(\cref{lemma:spectral})} \\
    & \leq \Big(1+\prod_{i=1}^d||J_i||_{2,2}\Big) \cdot \rho(s_1, s_2), \\
    & \leq \Big(1+\prod_{i=1}^d||J_i||_{2,2}\Big) \cdot \delta,
\end{align*}
which completes the proof.
\end{proof}

The robustness and generalization framework proposed by \cite{xu2012robustness} provides an approach to quantify the generalization error given that the algorithm satisfies the robustness defined by \ref{def:robust}. 

\begin{lemma} \label{lemma:xuetal}
\textit{(Adopted from Theorem 3 in \cite{xu2012robustness})} If $S$ consists of n i.i.d. samples, and $\mathcal{A}$ is $(K, \epsilon(S))$-robust, then for any $\delta > 0$, with probability at least $1-\delta$,
\begin{displaymath}
\Big|\hat l(\mathcal{A}_s)-l(\mathcal{A}_s)\Big| \leq \epsilon(S) + M \sqrt{\frac{2K\ln2 + 2\ln (1/\delta)}{n}},
\end{displaymath}
where $M$ is the maximum training loss.
\end{lemma}

Since we already show that $G_\theta$ is $\left(\mathcal{N}(\frac{\delta}{2};S,\rho), (1+\prod_{i=1}^{d}||J_i||_{2,2})\delta \right)$-robust, by following \cref{lemma:xuetal}, we conclude that:
\begin{displaymath}
\Big|\hat l(G_\theta)-l(G_\theta)\Big| \leq (1+\prod_{i=1}^{d}||J_i||_{2,2})\delta +  M \sqrt{\frac{2\mathcal{N}(\frac{\delta}{2};S,\rho)\ln2 + 2\ln (1/\delta)}{n}}.
\end{displaymath}

By incorporating the forward modeling of FWI, specifically, we embed the $L-lipschitz$ continuity and the upper bound of the noise imposed on the observed data. 

\begin{lemma}
Suppose $X \subset \mathbb{R}^p$ and $Y \subset \mathbb{R}^q$, the forward mapping $f: X \longrightarrow Y$ is \textit{L-lipschitz}. Denoting $S=X\times Y$ as the sample space equipped with metric $\rho$, we have:
\begin{displaymath}
\mathcal{N}(\frac{L\delta +2\eta}{2}; S, \rho) \leq \mathcal{N}(\frac{\delta}{2};X, l_2).
\end{displaymath}
\end{lemma}
\begin{proof}
Let $X'$ be a $\delta$-cover of $X$, then $\forall x \in X, \exists x' \in X'$ s.t. $||x-x'||_2 \leq \delta$. \par Let $Y'$ be the set that $Y':= \{y'=\mathcal{F}(x')+n, x'\in X', ||n'||_{2} \leq \eta \}$, and similarly $Y:= \{y=\mathcal{F}(x)+n, x\in X, ||n||_{2} \leq \eta \}$. It suffices to show that $S' = X' \times Y'$ is a $(L\delta + 2\eta)$-cover of $S$.  \par
For any $y \in Y$ and $y' \in Y'$, we have: 
\begin{align*}
    ||y-y'|| &\leq ||\mathcal{F}(x)+n-\mathcal{F}(x')-n'||_2 \\
    &\leq ||\mathcal{F}(x)-\mathcal{F}(x')||_2+||n-n'||_2 \\
    & \leq L||x-x'||_2 + 2\eta \\
    & \leq L\delta + 2\eta
\end{align*}
This indicates that $Y'$ is an $(L\delta + 2\eta)$-cover of Y. Now consider $\forall s\in S and s'\in S'$, and recall the definition of metric $\rho$, it follows that:
\begin{align*}
    \rho(s,s') &= \max (||x-x'||_2, ||y-y'||_2) \\
    &\leq \max(\delta, L\delta+2\eta) = L\delta+2\eta
\end{align*}
Therefore $S'$ is a $(L\delta+2\eta)$-cover of $S$, concluding the proof.
\end{proof}


\section{Experiments}
\label{sec:experiments}

\begin{figure}
    \centering
    \includegraphics[width=0.95\textwidth]{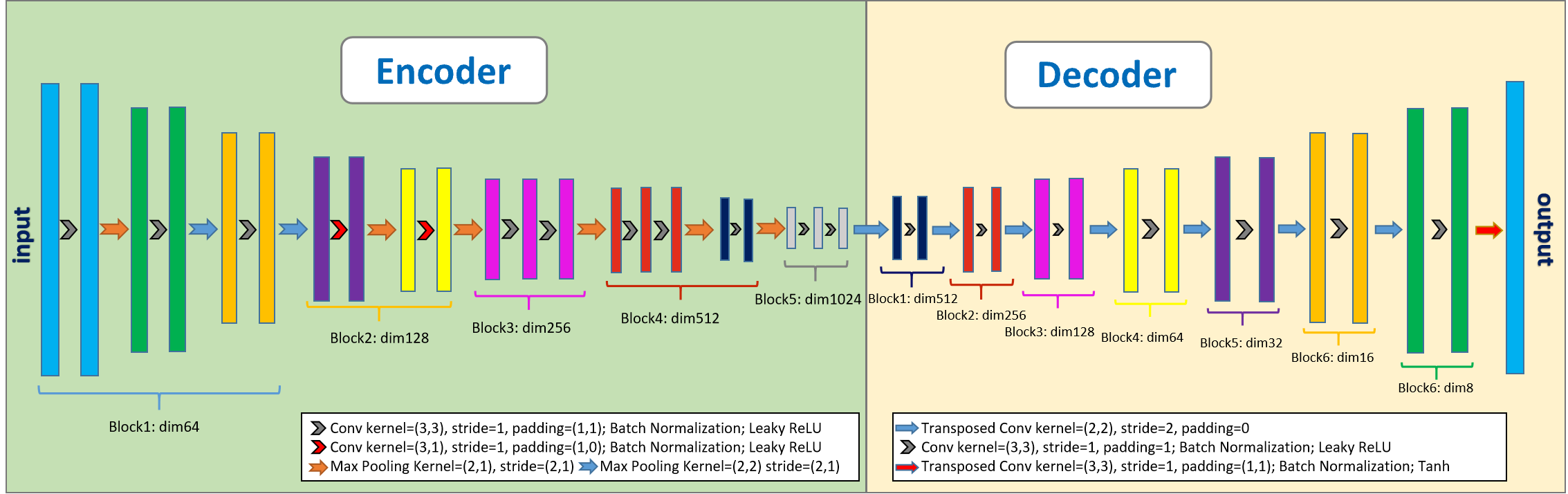}
    \caption{Model Architecture of the Encoder-decoder InversionNet used in this paper.}
    \label{fig:invnet}
\end{figure}

In this section, we show experimental results to provide interpretation on the proved bounds. Following the common experimental settings in the previous work\cite{wu2019inversionnet,yang2019deep}, we introduce three standard seismic FWI datasets published in OpenFWI~\cite{deng2021openfwi}: Kimberlina CO$_2$, Kimberlina Reservoir and the CurvedVel datasets. Both Kimberlina CO$_2$ and Reservoir datasets are created by the U.S. Department of Energy~(DOE). Specifically, the Kimberlina CO$_2$ dataset is to study and simulate CO$_2$ leakage through the well bore in the shallow layers~\cite{Characterizing-2017-DOE}, while CO$_2$ Reservoir dataset focuses on the migration of super-critical CO$_2$ after injecting into the storage reservoir~\cite{Development-2021-Alumbaugh}. We provide a visualization of a few samples and their corresponding shot-gather seismic data (\cref{fig:co2} and \cref{fig:reservoir})  in Appendix~D. CurvedVel dataset was generated to simulate curved subsurface layers with geologic fault zones and varying central frequencies. We provide a visualization of  several seismic shot-gathers generated with different central frequencies~(\cref{fig:cur_fre}) and a few velocity maps selected from CurvedVel dataset with different geologic faults~(\cref{fig:cur_fault}) in Appendix~E. We use the two Kimberlina datasets for robustness test and the CurvedVel dataset for generalization test. The statistics of three datasets are summarized in \cref{tab:dataset}. ``Data'' refers to the input seismic data, and ``label'' stands for the output velocity maps. As for the choice of Deep Neural Network, we implement an encoder-decoder based Convolutional Neural Network following the InversionNet~\cite{wu2019inversionnet}. A general model architecture is given in \cref{fig:invnet}. Note that the architecture and parameters vary with different datasets, and we provide a list of all details for each task in the supplementary materials.\par

\begin{table}[]
    \centering
    \begin{tabular}{ccccc}
    \specialrule{.15em}{.05em}{.05em} 
        \textbf{Dataset} &  \textbf{Train Size} & \textbf{Test Size} & \textbf{Data Shape} & \textbf{Label Shape} \\
         \specialrule{.1em}{.05em}{.05em} 
        Kimberlina CO$_2$ & 15000 & 4430 & 9$\times$1251$\times$101 & 401$\times$141 \\
        Kimberlina Reservoir & 1200 & 337 & 6$\times$ 1001$\times$ 200 & 601 $\times$ 351\\
        CurvedVel & 60000 & 7000 & 6$\times$200$\times$200 & 200$\times$200 \\
    \specialrule{.15em}{.05em}{.05em} 
    \end{tabular}
    \caption{Summary of the Datasets}
    \label{tab:dataset}
\end{table}

\subsection{Robustness Test}
As \cref{thm:robust} indicates, we target the upper bound of the deviation between the prediction with clean data and perturbed data, namely the performance drop,  as the robustness of the DNN model. Recall that \cref{cor:cond} states that the DNNs with MSE loss has a looser generalization bound than DNNs with MAE loss. Therefore, we trained the InversionNet shown in \cref{fig:invnet} on two Kimberlina Datasets~(CO$_2$ and Reservoir) with MAE loss and MSE loss, respectively. A detailed description and illustration of these two datasets are given in the Appendix. In the test phase, we impose noise to the test data with different levels of signal-to-noise ratios (SNR=\{inf, 30, 20, 10, 0\}, where inf stands for no noise). 

\begin{figure}
    \centering
    \includegraphics[width=0.95\textwidth]{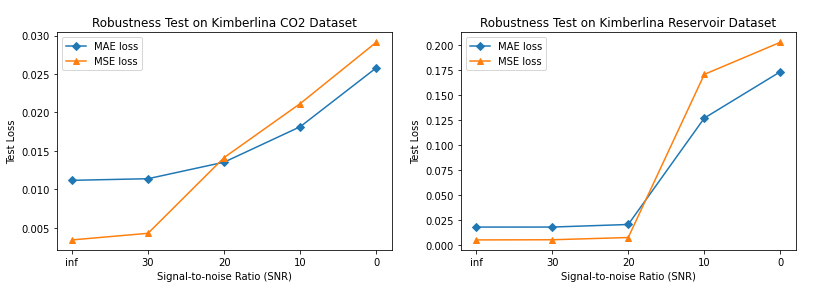}
    \caption{Both plots show similar dynamics between the test loss and SNR. Upon converge, the value of MSE loss is asymptotically the sqaure of MAE loss, as the noise increases (SNR diminishes), the gain on MSE loss is more significant than MAE loss.}
    \label{fig:rob_test}
\end{figure}

\cref{fig:rob_test} illustrates two types of prediction loss for all noise levels on two datasets mentioned above. It can be observed that when $SNR=inf$, the value of MSE loss is asymptotically the square of MAE loss, which means that the predicted images show similarly good quality. As the noise level increases, the value of MSE loss surges and surpasses the value of MAE loss when SNR is around 10.\par

It may be argued that the \textit{value of loss} does not absolutely capture the \textit{real quality} of the generated images. To tackle with this issue, we leveraged the well received structural similarity index~\cite{wang2004image} as a reference, which gives a perceptual evaluation on the similarity between the predicted and ground-truth images. The value of SSIM is normalized into $[0,1]$, where a larger value indicates a higher level of similarity. \cref{tab:rob_kim} gives a detailed demonstration of the robustness test with percentage of loss gain and the SSIM index. It can be observed that as the noise level increases, the percentage of loss gain differs vastly between two types of DNNs. The SSIM value of DNNs with MSE quickly drops close to 0.8 and the other remains around 0.9, also implying that the DNNs with MAE loss are able to produce velocity maps with relatively good quality. These experimental results corroborate with our theoretical results in \cref{thm:robust} and \cref{cor:cond}.

\begin{table}[]
    \centering
    \begin{tabular}{|c|c|c|c|c|c|}
         \specialrule{.1em}{.05em}{.05em} 
          &  No Noise (snr=inf) & snr=30 & snr=20 & snr=10 & snr=0\\
         \specialrule{.1em}{.05em}{.05em} 
         MAE loss & 0.01117 & 0.01138 & 0.01353 & 0.01814 & 0.02580 \\
         \hline
         \% of loss gain & - & 0.0188 & 0.02112 & 0.624 & 1.31 \\
         \hline
         SSIM (MAE) & 0.968 & 0.967 & 0.963 & 0.958 & 0.913\\
         \specialrule{.2em}{.05em}{.05em} 
         MSE loss & 0.00342 & 0.00428 & 0.01413 & 0.02115 & 0.02913 \\
         \hline
         \% of loss gain & - & 0.22 & 3.06 & 5.18 & 7.37 \\
         \hline
         SSIM (MSE) & 0.947 & 0.939 & 0.896 & 0.815 & 0.828\\
         \specialrule{.1em}{.05em}{.05em} 
    \end{tabular}
    \caption{Robustness Test on Kimberlina CO$_2$ Dataset}
    \label{tab:rob_kim}
\end{table}

\subsection{Generalization Test}
Recall that our results in \cref{thm:gen} indicates that the generalization bounds can be characterized by: the maximum training loss $L_m$, training dataset size $N$, and the product of weight matrices norm, denoted as $W_0 = \prod_{i=1}^{d}||W_i||_F$. In this section, we illustrate substantial empirical results of the interplay between the generalization gap and those parameters. Notice that in practice, the number of training samples often yield a direct impact on the model performance, hence we put together $\frac{L_m}{\sqrt{N}}$ as one parameter of the generalization bounds in \cref{thm:gen}.  To accommodate the training dataset size, we choose the CurvedVel dataset which contains 60,000 images for training and 7,000 for testing. \par
We first study the empirical generalization behavior in a standard setting: data samples are independent and identically distributed (\textit{i.i.d}) in both training and testing sets, where the generalization gap $Err_G$ is the difference of training and testing performances. We train a similar encoder-decoder Convolutional Neural Network as \cref{fig:invnet} with three different optimizers: SGD, AdaGrad and AdamW. The results illustrated in \cref{tab:tsize} and \cref{tab:weight_norm} are the average of all trained models. We can observe from \cref{tab:tsize} and \cref{tab:weight_norm} that both of the parameters: $\frac{L_m}{\sqrt{N}}$ and $W_0$ show a positive relationship with the generalization gap. \par

\begin{table}
    \centering
    \begin{tabular}{|c|c|c|c|c|c|}
     \specialrule{.1em}{.05em}{.05em} 
        $\frac{L_m}{N} (\times 10^6)$ & $\frac{0.7025}{\sqrt{20k}}\approx 497$ & $\frac{0.6827}{\sqrt{30k}}\approx 394$ & $\frac{0.7042}{\sqrt{40k}}\approx 352$ &
        $\frac{0.6665}{\sqrt{50k}}\approx 298$ & $\frac{0.6783}{\sqrt{60k}}\approx 277$ \\
        \hline
        $Err_G$ & 0.045 & 0.038 & 0.036 & 0.031 & 0.030 \\
    \specialrule{.1em}{.05em}{.05em} 
    \end{tabular}
    \caption{Generalization Gap with Different Training Size and Max Training Loss}
    \label{tab:tsize}
\end{table}

\begin{table}[]
    \centering
    \begin{tabular}{|c|c|c|c|}
     \specialrule{.1em}{.05em}{.05em} 
        $\prod_{i=1}^{d}||W_i||_F$ & $2.097e+11$ & $5.353e+12$ & $8.491e+10$ \\
        \hline
        $Err_G$ & 0.03225 & 0.03538 & 0.02856 \\
    \specialrule{.1em}{.05em}{.05em} 
    \end{tabular}
    \caption{Generalization Gap with Different Product of Weight Matrices Norm}
    \label{tab:weight_norm}
\end{table}

It is necessary to remark that in experiments for \cref{tab:tsize}, we are not able to fix another important factor $W_0$. Although a number of works have shown that the parameters deviate very little in large-scale neural networks upon convergence~\cite{long2019generalization}, it is still not rigorous to ignore the impact of $W_0$. As shown in \cref{tab:tsize}, we only claim that such a positive relationship can be observed. In experiments for \cref{tab:weight_norm}, we fix $N$ and change the neural network architecture so that more dynamics of $W_0$ can be brought in.

\subsubsection{Generalization Test with Distribution Drift}
The issue of distribution drift on testing data emerges as a common hurdle in the field study of FWI, as the model is usually trained on synthetic data which fixates some geological features~(such as source location, central frequency, etc). Inspired by such a practical concern, we also conduct an ablation study to show how our generalization bounds work in those scenarios. Specifically, we consider the distribution drift with the change of two geological features: number of geologic faults and the central frequency. We provide samples of the velocity maps and seismic data with the various geological features mentioned above in the Appendix. \par
The setup here is that we use the models trained with original dataset, which is 1-fault and generated with the central frequency $f=15~Hz$ by default, but test them with data has multiple-fault or generated with different central frequencies. For simplification, we combine the two parameters in \cref{tab:tsize} and \cref{tab:weight_norm} as $\frac{L_{m}}{N}\cdot \prod_{i=1}^{d}||W_i||_F$ and show the development of generalization gap $Err_{G}$. \cref{tab:fault} and \cref{tab:frequency} demonstrate the experimental results on test data with two geological features, respectively.\par

\begin{table}[]
    \centering
    \caption{Generalization Gap with Different Number of Faults}
    \begin{tabular}{|c|c|c|c|c|}\hline
    
       Parameter & 1-fault & 2-fault & 3-fault & 4-fault\\\hline
        $\frac{L_{m}}{N}\cdot \prod_{i=1}^{d}||W_i||_F=8.5e+8$  &0.02856  &0.03477  &0.06092 &0.08848 \\
        $\frac{L_{m}}{N}\cdot \prod_{i=1}^{d}||W_i||_F=1.0e+9$ &0.03225  &0.04058  &0.06709 &0.09362  \\
        $\frac{L_{m}}{N}\cdot \prod_{i=1}^{d}||W_i||_F=1.4e+9$ & 0.03417 &0.05242  &0.07481 &0.09701 \\
        $\frac{L_{m}}{N}\cdot \prod_{i=1}^{d}||W_i||_F=2.1e+9$  &0.03538  &0.05595  &0.08271 &0.1076 \\
        $\frac{L_{m}}{N}\cdot \prod_{i=1}^{d}||W_i||_F=2.2e+9$ & 0.03469 &0.05136  &0.07891 &0.1059 \\
                \hline
    \end{tabular}

    \label{tab:fault}
\end{table}

\begin{table}[]
    \centering
    \caption{Generalization Gap with Different Wave Frequencies}
    \begin{tabular}{|c|c|c|c|c|c|}\hline
    
       Parameter  & f=15Hz & f=20Hz & f=25Hz\\\hline
        $\frac{L_{m}}{N}\cdot \prod_{i=1}^{d}||W_i||_F=8.5e+8$  &0.02856 & 0.1799 & 0.3197 \\
        $\frac{L_{m}}{N}\cdot \prod_{i=1}^{d}||W_i||_F=1.0e+9$  &0.03225 & 0.2020 &0.3459 \\
        $\frac{L_{m}}{N}\cdot \prod_{i=1}^{d}||W_i||_F=1.4e+9$  &0.03417 & 0.2336 & 0.3671\\
        $\frac{L_{m}}{N}\cdot \prod_{i=1}^{d}||W_i||_F=2.1e+9$  &0.03538 & 0.2907 & 0.3852 \\
        $\frac{L_{m}}{N}\cdot \prod_{i=1}^{d}||W_i||_F=2.2e+9$  &0.03469 & 0.2818 & 0.3955\\
        \hline
    \end{tabular}

    \label{tab:frequency}
\end{table}

An immediate observation from \cref{tab:fault} is that for each model, the generalization gap $Err_G$ increases as the number of fault rises. This is due to the fact that as the increasing of the number of the geological faults in test data indicates a gain of distribution drift. If we focus on each column of multi-fault, as $\frac{L_{m}}{N}\cdot \prod_{i=1}^{d}||W_i||_F$ increases from $7.5e+8$ to $2.2e+9$, the generalization gap keeps increasing except for the last two rows, whose results are very close, thus leading to the conclusion as \cref{tab:weight_norm} does for 1-fault, which is the default test data.\par

\cref{tab:frequency} presents the generalization error for test data generated with different central frequencies among $f = \{15, 20, 25\}\,Hz$. Similarly, the more deviated from central frequency of 15\,Hz, the larger generalization gap can be observed. Again, fixing each column of a specific central frequency, a positive relationship between  $\frac{L_{m}}{N}\cdot \prod_{i=1}^{d}||W_i||_F$ and the generalization gap still holds as discussed in \cref{tab:weight_norm} and \cref{tab:fault}.


\section{Conclusion}
\label{sec:conclusion}
In this paper, motivated by the recent progress of utilizing deep neural networks for Full Waveform Inversion, we study from the theoretical perspectives on how these model perform against perturbation and generalize in practical scenarios. We consider the performance degradation when tested with perturbed data by upper-bounding the gain of loss, and prove that DNNs trained with MAE loss are more robust than DNNs trained with MSE loss as the noise increases. We also prove an upper bound for the generalization gap based on the norm of weight matrices. Experimental results are illustrated with standard FWI datasets on both topics, corroborating with our theoretical results. We further remark that the analysis can be extended to other deep learning driven inverse problems, paving the way of a better understanding of the exploit of deep learning in scientific domains. 

\section*{Acknowledgment}
This work was supported by the Center for Space and Earth Science at Los Alamos National Laboratory~(LANL), and by the Laboratory Directed Research and Development program under the project number 20210542MFR at LANL.

\clearpage
\appendix

\section{Proof of \cref{fact:mse}} \label{apdx:mse}
\begin{fact}
For $\forall x_1, x_2 \in \mathbb{R}^d$, and $\mathcal{L}(x, \hat x) = ||x-\hat x||_2$, $|\mathcal{L}(x_1, \hat x)-\mathcal{L}(x_2, \hat x)| \leq \mathcal{L}(x_1, x_2)$ does not hold for $\forall x \in \mathbb{R}^d$
\end{fact}
\begin{proof}
Let $x=2x_1$. We will have
\begin{align*}
    |\mathcal{L}(x_1, \hat x)-\mathcal{L}(x_2, \hat x)| &= |||x_1-x||_2^2-||x_2-x||_2^2|, \\
    &= |||x_1||_2^2-2x_1^Tx-||x_2||_2^2+2x_2^Tx|, \\
    &= |3||x_1||_2^2-4x_1^Tx_2+||x_2||_2^2|.
\end{align*}
Notice that $\mathcal{L}(x_1,x_2) = ||x_1-x_2||_2^2 = ||x_1||_2^2 -2x_1^Tx_2+||x_2||_2^2$. Without loss of generality, we further assume $||x_1||_2^2 > ||x_2||_2^2$, and it follows that:
\begin{align*}
    &\mathcal{L}(x_1,x_2)- |\mathcal{L}(x_1, \hat x)-\mathcal{L}(x_2, \hat x)| \\
    &= 2||x_1||_2^2-2x_1^{T}x_2, \\
    &\geq ||x_1||_2^2-2x_1^{T}x_2+||x_2||^2_2 \geq 0,
\end{align*}
which completes the proof.
\end{proof}

\section{Example of \cref{fact:conv}}
Supposing the convolution filter $c$ has dimension $d_c=2$ and the input image $x \in \mathbb{R}^{3 \times 3}$, we will have:
\begin{equation*}
    w = 
    \begin{bmatrix}
    w_{1,1} & w_{1,2}  \\
    w_{2,1} & w_{2,2} \\
    \end{bmatrix}
    ,  
    x = 
    \begin{bmatrix}
    x_{1,1} & x_{1,2} & x_{1,3} \\
    x_{2,1} & x_{2,2} & x_{2,3} \\
    x_{3,1} & x_{3,2} & x_{3,3} \\
    \end{bmatrix}.
\end{equation*}

For the sake of simplicity, let padding and stride both be 0, then a standard convolution operation follows:
\begin{equation*}
    conv(x, W) = 
    \begin{bmatrix}
    \sum_{i=1}^{2}\sum_{j=1}{2}w_{i,j}x_{i,j} & \sum_{i=1}^{2}\sum_{j=1}{2}w_{i,j}x_{i,j}\\
    \sum_{i=1}^{2}\sum_{j=1}{2}w_{i,j}x_{i,j} &
    \sum_{i=1}^{2}\sum_{j=1}{2}w_{i,j}x_{i,j}
    \end{bmatrix}.
\end{equation*}
If we reshape $x$ to dimension 1 such that $x\in \mathbb{R}^{9 \times 1}$, apparently $conv(x, W) = \mathrm{op} \cdot x$, where $\mathrm{op} \in \mathbb{R}^{4 \times 9}$, given explicitly as
\begin{equation*}
    \mathrm{op} = 
    \begin{bmatrix}
    w_{1,1} & w_{1,2} & 0 & w_{2,1} & w_{2,2} & 0 & 0 & 0 & 0\\
    0 & w_{1,1} & w_{1,2} & 0 & w_{2,1} & w_{2,2} & 0 & 0 & 0\\
    0 & 0 & 0 & w_{1,1} & w_{1,2} & 0 & w_{2,1} & w_{2,2} & 0\\
    0 & 0 & 0 & 0 & w_{1,1} & w_{1,2} & 0 & w_{2,1} & w_{2,2}
    \end{bmatrix}.
\end{equation*}

\section{Example of \cref{fact:deconv}}
\begin{figure}
    \centering
    \includegraphics[width=15cm]{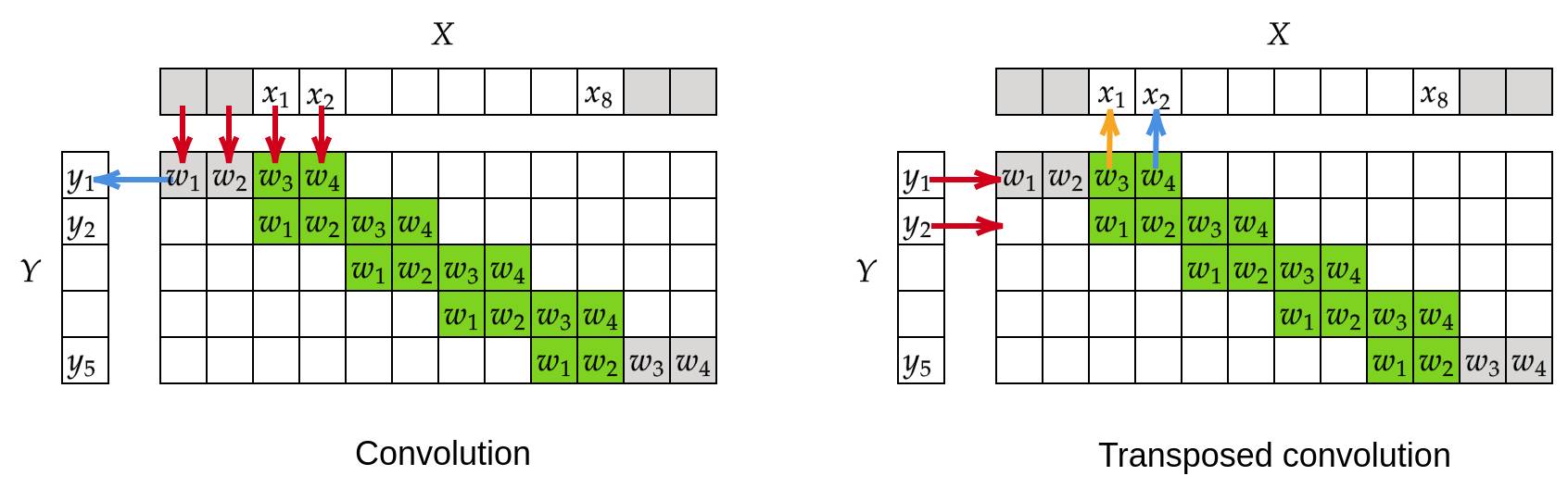}
    \caption{Illustration of Convolution and Transposed Convolution}
    \label{fig:conv_deconv}
\end{figure}

We illustrate the process of convolution and transposed convolution with 1D signals. As a general setting, the convolution filter with dimension 2 down-samples $X = (x_1, \dots, x_8)$ to $Y = (y_1, \dots, y_5)$, with both stride and padding equals 2. The weight of the filter is $w_i, i = \{1,2,3,4\}$. The transposed convolution, also termed as \textit{de-convolution}, takes $Y$ as input and reconstruct $X$ through up-sampling. \cref{fig:conv_deconv} provides an explicit illustration of both operations. For \cref{fact:deconv}, all we want to argue is that there also exists a matrix $\mathrm{op}$ such that the output of a transposed convolutional layer is also a linear transformation of the input features. This is obvious from \cref{fig:conv_deconv}.

\section{The Kimberlina CO$_2$ and Reservoir Datasets}
\begin{figure}
    \centering
    \includegraphics[width=15cm]{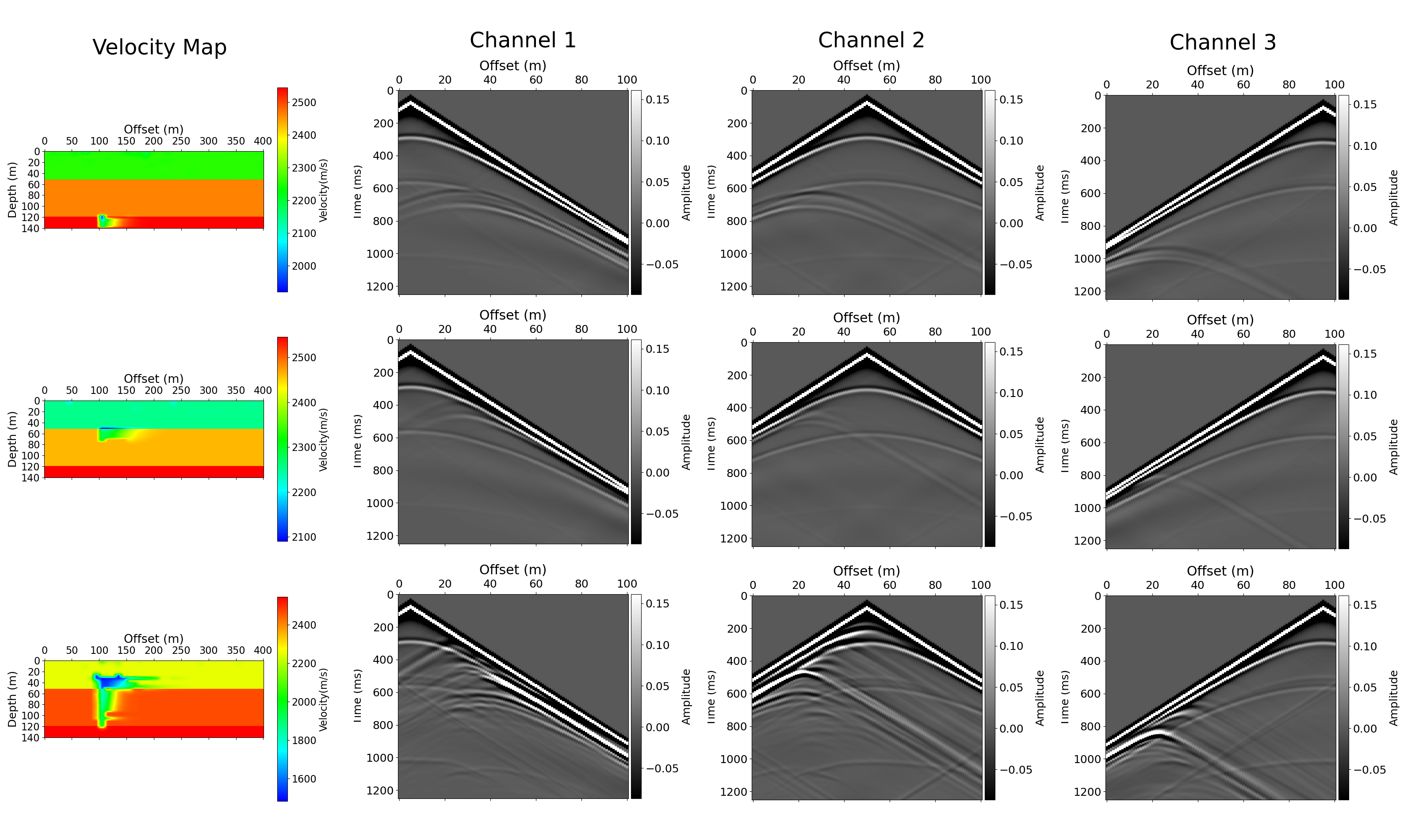}
    \caption{The Kimberlina CO$_2$ Dataset}
    \label{fig:co2}
\end{figure}

\begin{figure}
    \centering
    \includegraphics[width=15cm]{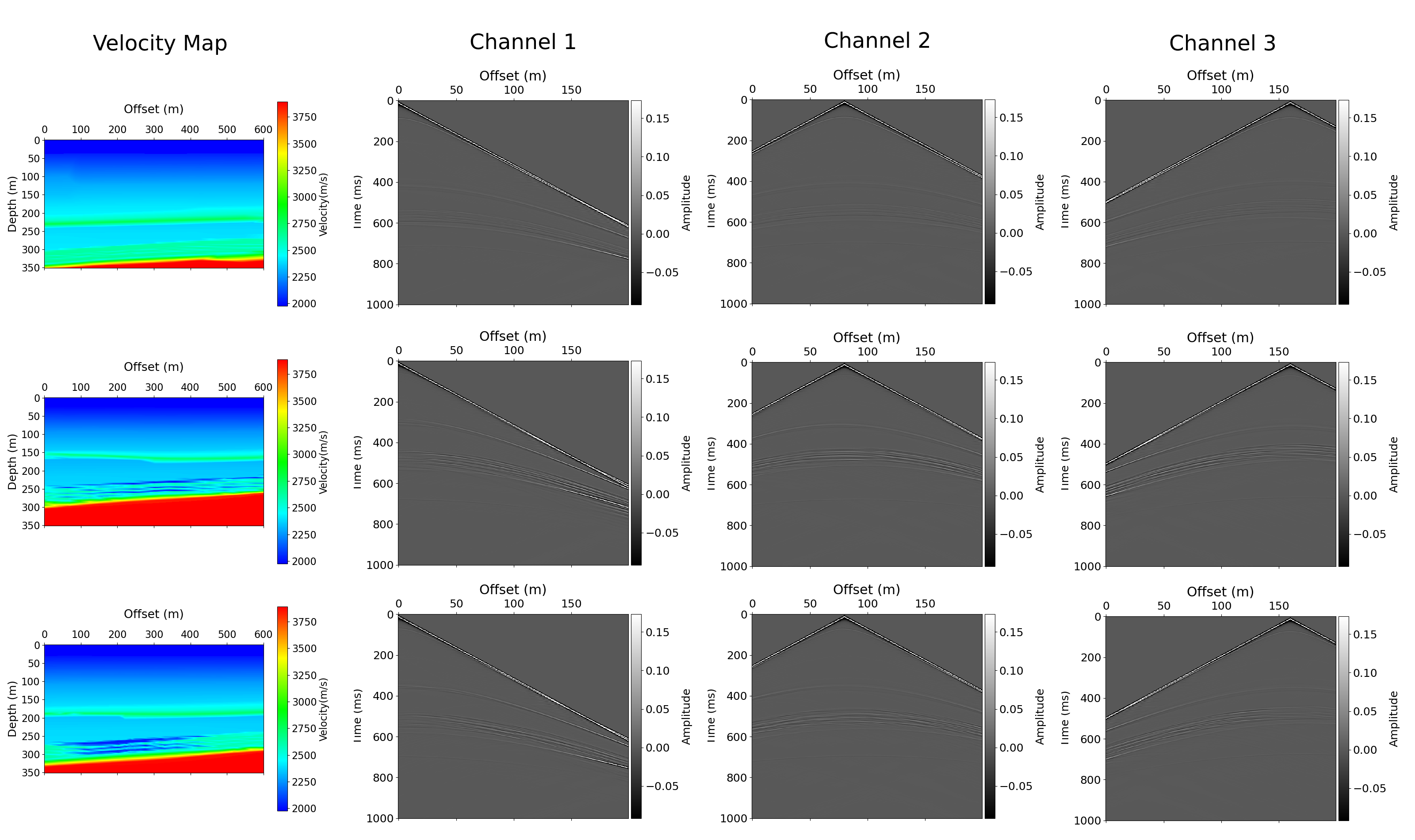}
    \caption{The Kimberlina Reservoir Dataset}
    \label{fig:reservoir}
\end{figure}

We provide illustrations of the Kimberlina CO$_2$ and Reservoir dataset. As shown in \cref{fig:co2} and \cref{fig:reservoir}, the first column presents the velocity maps and the rest columns are three channels of the seismic data w.r.t. different sources. It's worth mentioning that the Kimberlina $CO_2$ simulates the $CO_2$ leakage over a duration of 200 years, and the leakage mass can be categorized into: small, medium and large size, demonstrated as three rows in \cref{fig:co2}. 

\section{The CurvedVel Dataset}
\begin{figure}
    \centering
    \includegraphics[width=15cm]{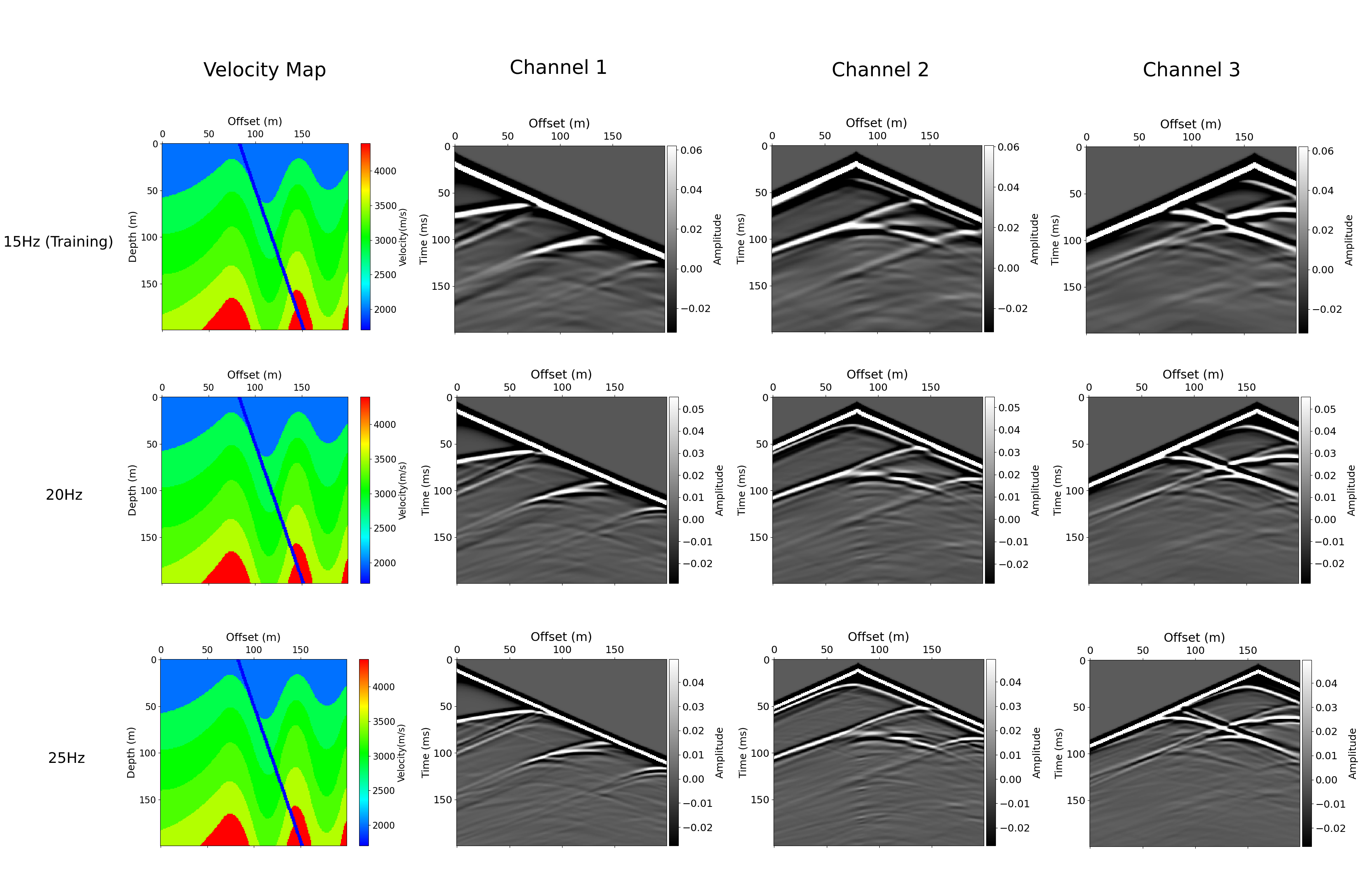}
    \caption{Illustration of CurvedVel dataset with different central frequencies}
    \label{fig:cur_fre}
\end{figure}

\begin{figure}
    \centering
    \includegraphics[width=15cm]{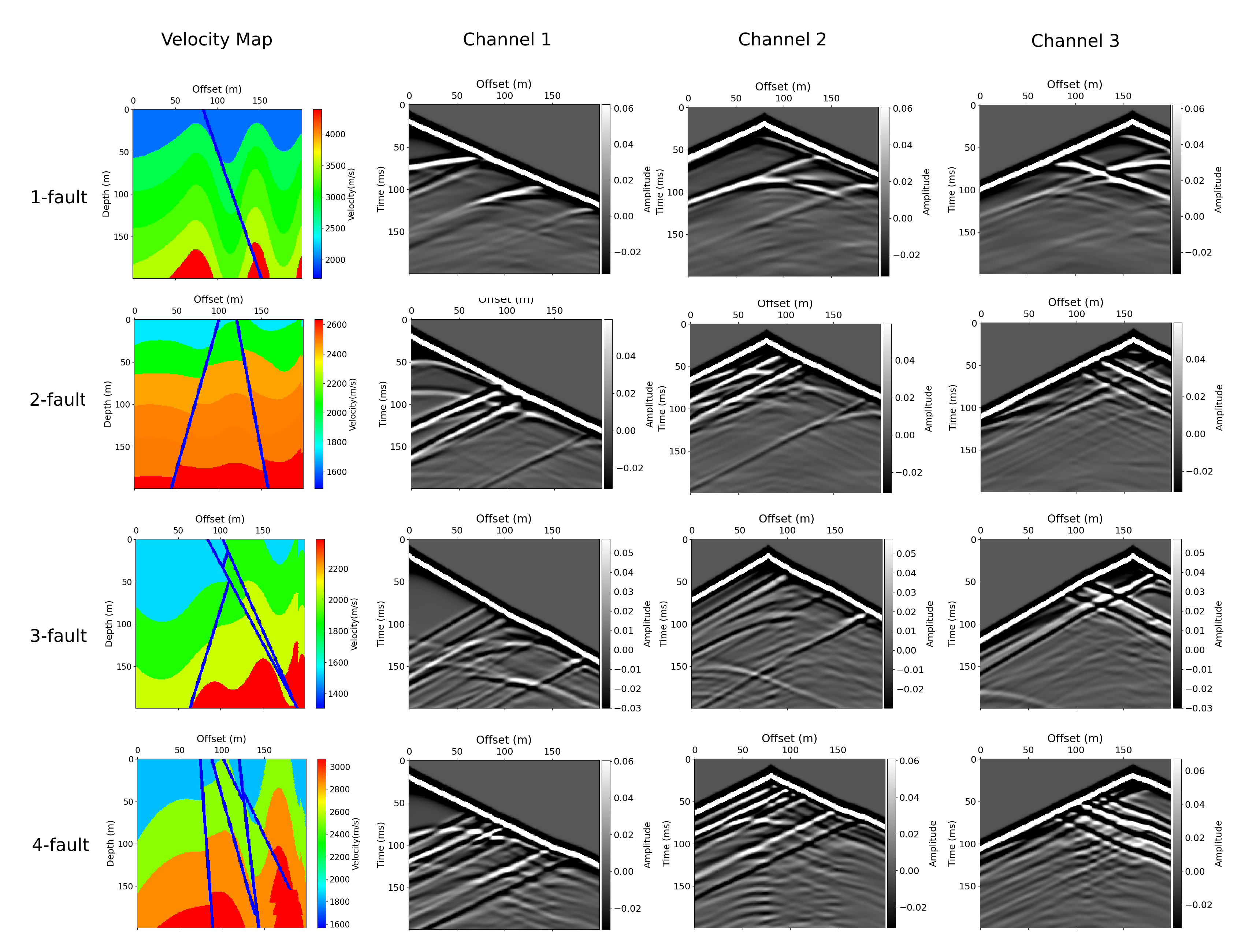}
    \caption{Illustration of CurvedVel dataset with multiple geologic faults}
    \label{fig:cur_fault}
\end{figure}

Recall that the CurvedVel dataset was synthesized with one geological fault and central frequency of 15 Hz, as shown in the first row in \cref{fig:cur_fre}. The rest rows in \cref{fig:cur_fre} correspond to velocity maps and the generated seismic data with central frequencies of 20Hz and 25Hz, respectively. Similarly, \cref{fig:cur_fault} illustrates velocity maps with 1, 2,3 and 4 geologic faults and the corresponding shot-gather seismic data with different sources.

\newpage
\bibliographystyle{unsrt}  
\bibliography{references}  

\end{document}